\pdfoutput=1
\documentclass{article} 
\usepackage{times}
\usepackage[utf8]{inputenc} 
\usepackage[T1]{fontenc}    
\usepackage{hyperref}       
\usepackage{url}            
\usepackage{booktabs}       
\usepackage{amsfonts}       
\usepackage{nicefrac}       
\usepackage{microtype}      
\usepackage{geometry}
\usepackage{authblk}
\usepackage{latexsym}

\usepackage{enumerate}
\usepackage[inline]{enumitem}
\usepackage{amsmath,amssymb}
\usepackage{amsfonts,dsfont}
\usepackage{nicefrac}
\usepackage{microtype}
\usepackage{mathtools}

\usepackage{caption}
\usepackage{subcaption}
\usepackage{wrapfig}
\usepackage{enumitem}

\usepackage[normalem]{ulem}
\usepackage{amssymb}
\usepackage{multicol}
\usepackage{adjustbox}
\usepackage{multirow}
\usepackage{color}
\usepackage{xspace}
\usepackage{CJKutf8}
\usepackage{graphicx}
\usepackage[ruled,vlined]{algorithm2e}

\PassOptionsToPackage{numbers}{natbib}
\usepackage{natbib}

\def\BibTeX{{\rm B\kern-.05em{\sc i\kern-.025em b}\kern-.08emT\kern-.1667em\lower.7ex\hbox{E}\kern-.125emX}}

\begin{document}

\title{\textbf{Leveraging Extracted Model Adversaries \\for Improved Black Box Attacks}}

\author{Naveen Jafer Nizar$^1$, Ari Kobren$^2$}
\affil{\{naveen.jafer, ari.kobren\}@oracle.com}
\affil{Oracle Corporation$^1$, Oracle Labs$^2$}

\maketitle

\begin{abstract}
We present a method for adversarial input generation against black box models for reading comprehension based question answering.
Our approach is composed of two steps.
First, we approximate a \emph{victim} black box model via model extraction~\cite{krishna2020thieves}.
Second, we use our own white box method to generate input perturbations that cause the approximate model to fail.
These perturbed inputs are used against the victim.
In experiments we find that our method improves on the efficacy of the \textsc{AddAny}---a white box attack---performed on the approximate model by 25\% F1, and the \textsc{AddSent} attack---a black box attack---by 11\% F1~\cite{jia2017adversarialRC}.
\end{abstract}

\section{Introduction}
Machine learning models are ubiquitous in technologies that are used by billions of people every day. In part, this is due to the recent success of deep learning. Indeed, research in the last decade has demonstrated that the most effective deep models can match or even outperform humans on a variety of tasks \cite{Devlin2019BERTPO,xie2020}.

Despite their effectiveness, deep models are also known to make embarrassing errors. 
This is especially troublesome when those errors can be categorized as unsafe, e.g., racist, sexist, etc.~\cite{wallace2019-universal}.  This leads to the desire for methods to audit models for correctness, robustness and---above all else---safety, before deployment.

Unfortunately, it is difficult to precisely determine the set of inputs on which a deep model fails because deep models are complex, have a large number of parameters---usually in the billions---and are non-linear \cite{Radford2019LanguageMA}. In an initial attempt to automate the discovery of inputs on which these embarrassing failures occur, researchers developed a technique for making calculated perturbations to an image that are imperceptible to the human eye, but cause deep models to misclassify the image \cite{szegedy2014intriguing}. In addition to developing more effective techniques for creating \emph{adversarial inputs} for vision models  \cite{papernot2017practical}, subsequent research extends these ideas to new domains, such as natural language processing (NLP).

NLP poses unique challenges for adversarial input generation because: 
\begin{enumerate*}
\item natural language is discrete rather than continuous (as in the image domain); and
\item in NLP, an ``imperceptible perturbation" of a sentence is typically construed to mean a semantically similar sentence, which can be difficult to generate.
\end{enumerate*}

Nevertheless, the study of adversarial input generation for NLP models has recently flourished, with techniques being developed for a wide variety of tasks such as: text classification, textual entailment and question answering~\cite{jin2019robustbert,wallace2019-universal,li2020bertattack,jia2017adversarialRC}.

These new techniques can be coarsely categorized into two groups: \emph{white box attacks}, where the attacker has full knowledge of the \emph{victim} model---including its parameters---and \emph{black box attacks}, where the attacker only has access to the victim's predictions on specified inputs. 
Unsurprisingly, white box attacks tend to exhibit much greater efficacy than black box attacks. 

In this work, we develop a technique for black box adversarial input generation for the task of reading comprehension that employs a white box attack on an approximation of the victim. More specifically, our approach begins  with \emph{model extraction}, where we learn an approximation of the victim model~\cite{krishna2020thieves}; afterward, we run a modification of the \textsc{AddAny}~\cite{jia2017adversarialRC} attack on the model approximation. Our approach is inspired by the work of \citet{papernot2017practical} for images and can also be referred to as a \emph{Black box evasion attack} on the original model. 

Since the \textsc{AddAny} attack is run on an \emph{extracted} (i.e., approximate) model of the victim, our modification encourages the attack method to find inputs for which the extracted model's top-k responses are all incorrect, rather than only its top response---as in the original \textsc{AddAny} attack.
The result of our \textsc{AddAny} attack is a set of adversarial perturbations, which are then applied to induce failures in the victim model. Empirically, we demonstrate that our approach is more effective than \textsc{AddSent}, i.e., a black box method for adversarial input generation for reading comprehension~\cite{jia2017adversarialRC}. 
Crucially, we observe that our modification of \textsc{AddAny} makes the attacks produced more robust to the difference between the extracted and victim model. In particular, our black box approach causes the victim to fail 11\% more than \textsc{AddSent}. While we focus on reading comprehension, we believe that our approach of model extraction followed by white box attacks is a fertile and relatively unexplored area that can be applied to a wide range of tasks and domains.

\textbf{Ethical Implications:} The primary motivation of our work is helping developers test and probe models for weaknesses before deployment. While we recognize that our approach could be used for malicious purposes we believe that our methods can be used in an effort to promote model safety.

\section{Background}
In this section we briefly describe the task of \emph{reading comprehension based question answering}, which we study in this work. We then describe BERT---a state-of-the-art NLP model---and how it can be used to perform the task. 

\subsection{Question Answering}
One of the key goals of NLP research is the development of models for \emph{question answering} (QA). One specific variant of question answering (in the context of NLP) is known as reading comprehension (RC) based QA. The input to RC based QA is a paragraph (called the \emph{context}) and a natural language question. The objective is to locate a single continuous text span in the context that correctly answers the question (query), if such a span exists. 

\subsection{BERT for Question Answering}
A class of language models that have shown great promise for the RC based QA task are BERT (Bidirectional Encoder Representations from Transformers as introduced by \citet{Devlin2019BERTPO}) and its variants. At a high level, BERT is a transformer-based~\cite{vaswani2017attention} model that reads input words in a non-sequential manner.  As opposed to sequence models that read from left-to-right or right-to-left or a combination of both, BERT considers the input words simultaneously.

BERT is trained on two objectives: One called masked token prediction (MTP) and the other called next sentence prediction (NSP). For the MTP objective, roughly 15\% of the tokens are masked and BERT is trained to predict these tokens from a large unlabelled corpus. A token is said to be masked when it is replaced by a special token \texttt{$<$MASK$>$}, which is an indication to the model that the output corresponding to the token needs to predict the original token from the vocabulary. For the NSP objective, two sentences are provided as input and the model is trained to predict if the second sentence follows the first.~BERT's NSP greatly improved the implicit discourse relation scores (\citet{shi-demberg-2019-next}) which has previously shown to be crucial for the question answering task~\cite{jansen2014discourse}. 

Once the model is trained on these objectives, the core BERT layers (discarding the output layers of the pre-training tasks) are then trained further for a downstream task such as RC based QA. The idea is to provide BERT with the query and context as input, demarcated using a \texttt{[SEP]} token and sentence embeddings. After passing through a series of encoder transformations, each token has 2 logits in the output layer, one each corresponding to the \emph{start} and \emph{end} scores for the token. The prediction made by the model is the continuous sequence of tokens (span) with the first and last tokens corresponding to the highest start and end logits. Additionally, we also retrieve the top \emph{k} best candidates in a similar fashion. 
\section{Method}
Our goal is to develop an effective black box attack for RC based QA models. Our approach proceeds in two steps: first, we build an approximation of the victim model, and second, we attack the approximate model with a powerful white box method.~The result of the attack is a collection of adversarial inputs that can be applied to the victim.~In this section we describe these steps in detail.

\subsection{Model Extraction}
\label{sec:extract}
The first step in our approach is to build an approximation of the victim model via \emph{model extraction}~\cite{krishna2020thieves}. At a high level, this approach constructs a training set by generating inputs that are served to the victim model and collecting the victim's responses. The responses act as the labels of the inputs. After a sufficient number of inputs and their corresponding labels have been collected, a new model can be trained to predict the collected labels, thereby mimicking the victim. The approximate model is known as the \emph{extracted} model. 

The crux of model extraction is an effective method of generating inputs. 
Recall that in RC based QA, the input is composed of a query and a context. 
Like previous work, we employ 2 methods for generating contexts: \textsc{WIKI} and \textsc{RANDOM}~\cite{krishna2020thieves}.
In the \textsc{WIKI} scheme, contexts are randomly sampled paragraphs from the WikiText-103 dataset.
In the \textsc{RANDOM} scheme, contexts are generated by sampling random tokens from the WikiText-103 dataset. 
For both schemes, a corresponding query is generated by sampling random words from the context.
To make the queries resemble questions, tokens such as ``where," ``who," ``what," and ``why," are inserted at the beginning of each query, and a ``?" symbol is appended to the end.
Labels are collected by serving the sampled queries and contexts to the victim model.
Together, the queries, contexts, and labels are used to train the extracted model.
An example query-context pair appears in Table \ref{exampleExtraction}.

\subsection{Adversarial Attack}
\label{sec:length}

\begin{center}

\begin{figure}
\centering

  \includegraphics[width=0.7\linewidth]{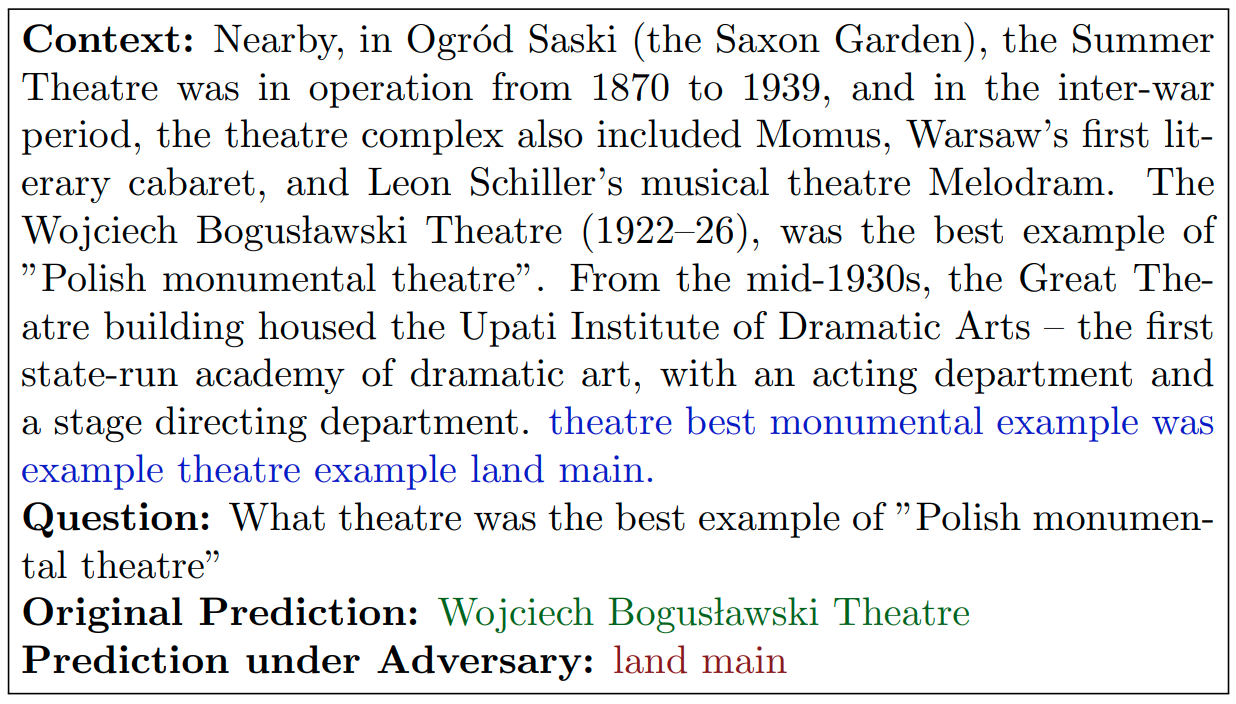}
  \caption{An example from SQuAD v1.1. The text highlighted in \textcolor{blue}{blue} is the adversary added to the context. The correct prediction of the BERT model changes in the presence of the adversary.}
  \label{fig:figure1}
\end{figure}
\end{center}

A successful adversarial attack on an RC base QA model is a modification to a context that preserves the correct answer but causes the model to return an incorrect span. We study \emph{non-targeted attacks}, in which eliciting any incorrect response from the model is a success (unlike \emph{targeted attacks}, which aim to elicit a \emph{specific} incorrect response form the model). Figure \ref{fig:figure1} depicts a successful attack. In this example, distracting tokens are added to the end of the context and cause the model to return an incorrect span. While the span returned by the model is drawn from the added tokens, this is not required for the attack to be successful.

\subsubsection{The \textsc{AddAny} Attack}
\label{sec:addany}
At a high level, the \textsc{AddAny} attack, proposed by \citet{jia2017adversarialRC}, generates adversarial examples for RC based QA models by appending a sequence of distracting tokens to the end of a context. 
The initial distracting tokens are iteratively exchanged for new tokens until model failure is induced, or a pre-specificed number of exchanges have been exceeded.
Since the sequence of tokens is often nonsensical (i.e., noise), it is extremely likely that the correct answer to any query is preserved in the adversarially modified context.

In detail, \textsc{AddAny} proceeds iteratively.
Let $q$ and $c$ be a query and context, respectively, and let $f$ 
be an RC based QA model whose inputs are $q$ and $c$ and whose output, $\mathcal{S} = f(c, q)$, is a distribution over token spans of $c$ (representing possible answers).
Let $s_i^\star = \arg\max \mathcal{S}_i$, i.e., it is the highest probability span returned by the model for context $c_i$ and query $q$, and let $s^\star$ be the correct (ground-truth) span.
The \textsc{AddAny} attack begins by appending a sequence of $d$ tokens (sampled uniformly at random) to $c$, to produce $c_1$.
For each appended token, $w_j$, a set of words, $W_j$, is initialized from a collection of common tokens and from tokens that appear in $q$.
During iteration $i$, compute $\mathcal{S}_i = f(c_i, q)$, and calculate the F1 score of $s_i^\star$ (using $s^\star$).
If the F1 score is 0, i.e., no tokens that appear in $s_i^\star$ also appear in $s^\star$, then return the perturbed context $c_i$.
Otherwise, for each appended token $w_j$ in $c_i$, iteratively exchange $w_j$ with each token in $W_j$ (holding all $w_k, k\ne j$ constant) and evaluate the \emph{expected} F1 score with respect to the corresponding distribution over token spans returned by $f$. 
Then, set $c_{i+1}$ to be the perturbation of $c_i$ with the smallest expected F1 score.
Terminate after a pre-specified number of iterations.
For further details, see \citet{jia2017adversarialRC}.

\subsubsection{\textsc{AddAny-kBest}}
\label{addanykBestSection}
During each iteration, the \textsc{AddAny} attack uses the victim model's distribution over token spans, $\mathcal{S}_i$, to guide construction of the adversarial sequence of tokens.
Unfortunately, this distribution is not available when the victim is a black box model.
To side-step this issue, we propose: \begin{enumerate*}[label=\roman*)]
    \item building an approximation of the victim, i.e., the extracted model (Section \ref{sec:extract}),
    \item for each $c$ and $q$, running \textsc{AddAny} on the extracted model to produce an adversarially perturbed context, $c_i$, and
    \item evaluating the victim on the perturbed context.
\end{enumerate*} 
The method succeeds if the perturbation causes a decrease in F1, i.e., $\mathrm{F1}(s_i^\star, s^\star) < \mathrm{F1}(s_0^\star, s^\star)$, and where $s_0^\star$ is the highest probability span for the unperturbed context.

Since the extracted model is constructed to be similar to the victim, it is plausible for the two models to have similar failure modes.
However, due to inevitable differences between the two models, even if a perturbed context, $c_i$, induces failure in the extracted model, failure of the victim is not guaranteed. 
Moreover, the \textsc{AddAny} attack resembles a type of over-fitting: as soon as a perturbed context, $c_i$, causes the extracted model to return a span, $s_i^\star$ for which $\mathrm{F1}(s_i^\star, s^\star) = 0$, $c_i$ is returned.
In cases where $c_i$ is discovered via exploitation of an artifact of the extracted model that is not present in the victim, the approach will fail.

To avoid this brittleness, we present \textsc{AddAny-kBest}, a variant of \textsc{AddAny}, which constructs perturbations that are more robust to differences between the extracted and victim models. 
Our method is parameterized by an integer $k$.
Rather than terminating when the highest probability span returned by the extracted model, $s_i^\star$, has an F1 score of 0, \textsc{AddAny-kBest} terminates when the F1 score for \emph{all} of the $k$-best spans returned by the extracted model have an F1 score of 0 or after a pre-specified number of iterations.
Precisely, let $S_i^k$ be the $k$ highest probability token spans returned by the extracted model, then terminate when:
\begin{align*}
    \max_{s \in S_i^k} \mathrm{F1}(s, s^\star) = 0.
\end{align*}
If the $k$-best spans returned by the extracted model all have an F1 score of 0, then \emph{none} of the tokens in the correct (ground-truth) span appear in \emph{any} of the $k$-best token spans.
In other words, such a case indicates that the context perturbation has caused the extracted model to lose sufficient confidence in all spans that are at all close to the ground-truth span.
Intuitively, this method is more robust to differences between the extracted and victim models than \textsc{AddAny}, and explicitly avoids constructing perturbations that only lead to failure on the best span returned by the extracted model. 

Note that a \textsc{AddAny-kBest} attack may not discover a perturbation capable of yielding an F1 of 0 for the $k$-best spans within the pre-specified number of iterations.
In such situations, a perturbation is returned that minimizes the expected F1 score among the $k$-best spans.
We also emphasize that, during the \textsc{AddAny-kBest} attack, a perturbation may be discovered that leads to an F1 score of 0 for the best token span, but unlike \textsc{AddAny}, this does not necessarily terminate the attack.

\section{Experiments}
In this section we present results of our proposed approach. 
We begin by describing the dataset used, and then report on model extraction.
Finally, we compare the effectiveness of \textsc{AddAny-kBest} to 2 other black box approaches. 

\subsection{Datasets} 
For the evaluation of RC based QA we use the SQuAD dataset \citep{rajpurkar2016squad}. Though our method is applicable to both v1.1 and v2.0 versions of the dataset we only experiment with \textsc{AddAny} for SQuAD v1.1 similar to previous investigations. Following \citep{jia2017adversarialRC}, we evaluate all methods on 1000 queries sampled at random from the development set.
Like previous work, we use the Brown Common word list corpus~\citep{francis79browncorpus} for sampling the random tokens (Section \ref{sec:addany}).

\begin{table}
\centering
\begin{tabular}{lrr}
\hline
\textbf{Model}  & \textbf{F1} & \textbf{EM}\\
\hline
VICTIM & 89.9 & 81.8 \\
WIKI & 83.6 & 73.5   \\
RANDOM & 75.8 & 63.2 \\
\hline
\end{tabular}
\caption{
A comparison of the original model (VICTIM) against the extracted models generated using 2 different schemes(RANDOM and WIKI). bert-base-uncased has been used as the LM in all the models mentioned above. All the extracted models use the same number of queries (query budget of 1x) as in the SQuAD training set. We report on the F1 and EM (Exact Match) scores for the evaluation set (1000 questions) sampled from the dev dataset. 
}
\label{extractionTable}
\end{table}

\begin{table}
\centering
\begin{tabular}{lrr}
\hline
\textbf{Model} & \textbf{Original (F1)}  & \textbf{\textsc{AddAny} (F1)} \\
\hline
Match LSTM single & 71.4 & 7.6  \\
Match LSTM ensemble & 75.4 & 11.7  \\
BiDAF single & 75.5 & 4.8  \\
BiDAF ensemble & 80.0 & 2.7  \\
\textbf{bert-base-uncased} & \textbf{89.9} & \textbf{5.9}  \\
\hline
\end{tabular}
\caption{
A comparison of the results of Match LSTM, BiDAF as reported by \citet{jia2017adversarialRC} with the bert-base-uncased model for SQuAD 1.1. We follow the identical experimental setup. The results for Match LSTM and BiDAF models were reported for both the single and ensemble versions.
}
\label{jiaAddOnResults}
\end{table}
\subsection{Extraction}
First, we present results for \textsc{WIKI} and \textsc{RANDOM} extraction methods (Section \ref{sec:extract}) on SQuAD v1.1 using a bert-base-uncased model for both the victim and extracted model in Table  \ref{extractionTable}. 

\paragraph{Remarks on Squad v2.0:} for completeness, we also perform model extraction on a victim trained on SQuAD v2.0, but the extracted model achieves significantly lower F1 scores. In SQuAD v1.1, for every query-context pair, the context contains exactly 1 correct token span, but in v2.0, for 33.4\% of pairs, the context \emph{does not contain} a correct span. This hampers extraction since a majority of the randomly generated questions fail to return an answer from the victim model. The extracted WIKI model has an F1 score of 57.9, which is comparably much lower to the model extracted for v1.1. 

We believe that the F1 of the extracted model for SQuAD v2.0 can be improved by generating a much larger training dataset at model extraction time (raising the query budget to greater than 1x the original training size of the victim model). But by doing this, any comparison in our results with SQuAD v1.1 would not be equitable.

\subsection{Methods Compared}
We compare \textsc{AddAny-kBest} to 2 baseline, black-box attacks: \begin{enumerate*}[label=\roman*)]
    \item the standard \textsc{AddAny} attack on the extracted model, and
    \item \textsc{AddSent}~\cite{jia2017adversarialRC}.
\end{enumerate*}
Similar to \textsc{AddAny}, \textsc{AddSent}  generates adversaries by appending tokens to the end of a context. 
These tokens are taken, in part from the query, but are also likely to preserve the correct token span in the context. 
In more detail, \textsc{AddSent} proceeds as follows:

\begin{enumerate}
    \item A copy of the query is appended to the context, but nouns and adjectives are replaced by their antonyms, as defined by WordNet~\cite{miller1995}. Additionally, an attempt is made to replace every named entity and number with tokens of the same part-of-speech that are nearby with respect to the corresponding GloVe embeddings~\cite{Pennington14glove:global}. If no changes were made in this step, the attacks fails.
    \item Next, a spurious token span is generated with the same type (defined using NER and POS tags
    from Stanford CoreNLP~\cite{Manning14thestanford} as the correct token span. Types are hand curated using NER and POS tags and have associated fake answers.
    \item The modified query and spurious token span are combined into declarative form using hand crafted rules defined by the CoreNLP constituency parses.
    \item Since the automatically generated sentences could be unnatural or ungrammatical, crowd-sourced workers correct these sentences. (This final step is not performed in our evaluation of AddSent since we aim to compare other fully automatic methods against this).  
\end{enumerate}
Note that unlike \textsc{AddAny}, \textsc{AddSent} does not require access to the model's distribution over token spans, and thus, it does not require model extraction.

\textsc{AddSent} may return multiple candidate adversaries for a given query-context pair. 
In such cases, each candidate is applied and the most effective (in terms of reducing instance-level F1 of the victim) is used in computing overall F1. To represent cases without access to (many) black box model evaluations, \citet{jia2017adversarialRC} also experiment with using a randomly sampled candidate per instance when computing overall F1. This method is called \textsc{AddOneSent}

For the \textsc{AddAny} and \textsc{AddAny-kBest} approaches, we also distinguish between instances in which they are run on models extracted via the WIKI (\textsc{W-A-argMax}, \textsc{W-A-kBest})or RANDOM (\textsc{R-A-argMax}, \textsc{R-A-kBest}) approaches. 

We use the same experimental setup as \citet{jia2017adversarialRC}. Additionally we experiment while both prefixing and suffixing the adversarial sentence to the context. This does not result in drastically different F1 scores on the overall evaluation set. However, we did notice that in certain examples, for a given context $c$, the output of the model differs depending on whether the same adversary was being prefixed or suffixed. It was observed that sometimes prefixing resulted in a successful attack while suffixing would not and vice versa. Since this behaviour was not documented to be specifically favouring either suffixing or prefixing, we stick to suffixing the adversary to the context as done by \citet{jia2017adversarialRC}.

\begin{table}
\centering
\begin{tabular}{lrr}
\hline
\textbf{Method} & \textbf{Extracted (F1)}  & \textbf{Victim (F1)} \\
\hline
\textbf{W-A-kBest} & 10.9 & \textbf{42.4} \\
W-A-argMax & 9.7 & 68.3 \\
R-A-kBest & 3.6 & 52.2 \\
R-A-argMax & 3.7 & 76.1 \\ 
\hline
AddSent & - & 53.2 \\
AddOneSent & - & 56.5 \\ 
\hline
Combined & - & 31.9 \\
\hline

\end{tabular}

\caption{\label{citation-guide}
The first 4 rows report the results for experiments on variations of \textsc{AddAny} (kBest/argMax) and extraction schemes (WIKI and RANDOM). The ``extracted" column lists the F1 score of the respective method used for generating adversaries. The ``victim" column is the F1 score on the victim model when transferred from the extracted (for \textsc{AddAny} methods). For \textsc{AddSent} and  \textsc{AddOneSent} it is the F1 score when directly applied on the victim model. The last row ``Combined" refers to the joint coverage of \textsc{W-A-kBest} + \textsc{AddSent}.  
}
\label{addAnyresults}
\end{table}

\subsection{Results}
In Table \ref{addAnyresults}, we report the F1 scores of all methods on the extracted model. The results reveal that the \textsc{kBest} minimization (Section \ref{addanykBestSection}) approach is most effective at reducing the F1 score of the victim. Notably, we observe a difference of over 25\% in the F1 score between \textsc{kBest} and \textsc{argMax} in both \textsc{WIKI} and \textsc{RANDOM} schemes. 

Interestingly, the \textsc{AddSent} and \textsc{AddOneSent} attacks are more effective than the \textsc{AddAny-argMax} approach but less effective than the \textsc{AddAny-kBest} approach. In particular they reduce the F1 score to 53.2 (\textsc{AddSent}) and 56.5 (\textsc{AddOneSent}) as reported in Table \ref{addAnyresults}. For completeness, we compare the \textsc{AddAny} attack on the victim model (similar to the work done in \citet{jia2017adversarialRC} for LSTM and BiDAF models. Table \ref{jiaAddOnResults} shows the results for bert-base-uncased among others for SQuAD v1.1. Only \textsc{argMax} minimization is carried out here since there is no post-attack transfer.  

\begin{figure}
\centering
  \includegraphics[width=0.7\linewidth]{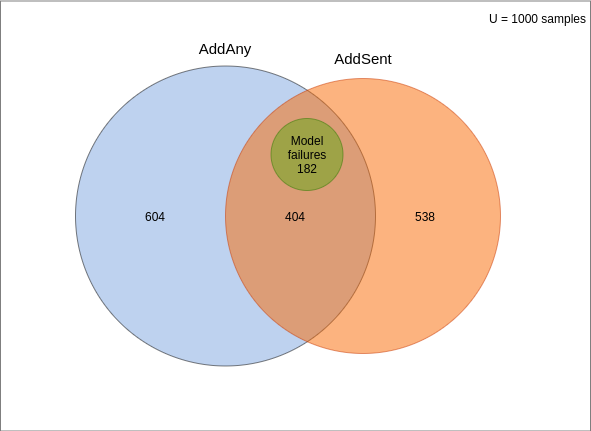}
  \caption{Joint coverage of \textsc{WIKI-AddAny-kBest} and \textsc{AddSent} on the evaluation}.
  \label{fig:vennAddSentAddAny}
\end{figure}

We also study the coverage of \textsc{W-A-kBest} and \textsc{AddSent} on the evaluation dataset of 1000 samples (Figure \ref{fig:vennAddSentAddAny}). \textsc{W-A-Kbest} and \textsc{AddSent} induce an F1 score of 0 on 606 and 538 query-context pairs, respectively. Among these failures, 404 query-context pairs were common to both the methods. Of the 404, 182 samples were a direct result of model failure of bert-base-uncased (exact match score is 81.8 which amounts to the 182 failure samples). If the methods are applied jointly, only 260 query-context pairs produce the correct answer corresponding to an exact match score of 26 and an F1 score of 31.9 (Table \ref{addAnyresults}). This is an indication that the 2 attacks in conjunction (represented by the ``Combined" row in Table \ref{addAnyresults}) provide wider coverage than either method alone.

\subsection{Fine-grained analysis}
In this section we analyze how successful the adversarial attack is for each answer \emph{category}, which were identified in previous work \cite{rajpurkar2016squad}. Table \ref{fineGrainedAnalysis} lists the 10 categories of ground-truth token spans, their frequency in the evaluation set as well as the average F1 scores on the victim model before and after the adversarial attack. We observe that ground-truth spans of type ``places" experienced a drastic drop in F1 score. ``Clauses" had the highest average length and also had the highest drop in F1 score subject to the \textsc{W-A-kBest} attack(almost double the average across classes). Category analysis such as this could help the community understand how to curate better attacks and ultimately train a model that is more robust on answer types that are most important or relevant for specific use cases.

\begin{table}
\centering
\begin{tabular}{lrrrr}
\hline
\textbf{Category} & \textbf{Freq \%} & \textbf{Before} & \textbf{After} & \textbf{Av-Len} \\ 
\hline
Names & 7.4 & 96.2 & 51.8 & 2.8\\
Numbers & 11.4 & 92.1 & 51.3 & 2\\
Places & 4.2 & 89 & 19.2 & 2.7\\
Dates & 8.4 & 96.8 & 40.3 & 2.1\\
Other Ents & 7.2 & 90.9 & 58.7 & 2.5\\
Noun Phrases & 48 & 88 & 41.8 & 2.3\\
Verb Phrases & 2.7 & 91.1 & 41.1 & 4.8\\
Adj Phrases & 1.9 & 70.3 & 27.8 & 1.6\\
Clauses & 1.3 & 82.9 & 7.6 & 6.8\\
Others & 9.5 & 89.7 & 34.7 & 5\\
\hline
\textbf{Total} & \textbf{100} & \textbf{89} & \textbf{42.4} & \textbf{2.7}\\
\hline

\end{tabular}

\caption{\label{citation-guide}
There are 10 general categories into which the answer spans have been classified. The first 4 are entities and \emph{Other Ents} is all other entities that do not fall into any of the 4 major categories. The 2nd column is the frequency of ground truth answers belonging to each of the categories. The 3rd column (Before) refers to the F1 score of questions corresponding to the category when evaluated on the Victim model. The 4th column (After) refers to the F1 score of questions corresponding to the category when evaluated on the Victim model under the presence of adversaries generated using \textsc{WIKI-AddAny-kBest} method. \emph{Av-Len} column is the average length of the answer spans in each category. 
}
\label{fineGrainedAnalysis}
\end{table}

\section{Related Work}
Our work studies black box adversarial input generation for reading comprehension. The primary building blocks of our proposed approach are model extraction, and white box adversarial input generation, which we discuss below. We also briefly describe related methods of generating adversarial attacks for NLP models.

A contemporary work that uses a similar approach to ours is \citet{wallace2020imitation}. While we carry out model extraction using non-sensical inputs, their work uses high quality out of distribution (OOD) sentences for extraction of a machine translation task. It is noteworthy to mention that in the extraction approach we follow~\cite{krishna2020thieves} the extracted model reaches within 95\% F1 score of the victim model with the same query budget that was used to train the victim model. This is in contrast to roughly 3x query budget taken in extracting the model in their work. The different nature of the task and methods followed while querying OOD datasets could be a possible explanation for the disparities.

\paragraph{Nonsensical Inputs and Model Extraction:} Nonsensical inputs to text-based systems have been the subject of recent study, but were not explored for extraction until recently \citep{krishna2020thieves}. \citet{feng2018-pathologies} studied model outputs while trimming down inputs to an extent where the input turned nonsensical for a human evaluator. Their work showed how nonsensical inputs produced overly confident model predictions. Using white box access to models \citet{wallace2019-universal} discovered that it was possible to generate input-agnostic nonsensical triggers that are effective adversaries on existing models on the SQuAD dataset. 

\paragraph{Adversarial attacks:} The first adversarial attacks against block box, deep neural network models focused on computer vision applications~\cite{papernot2017practical}. In concept, adversarial perturbations are transferable from computer vision to NLP; but, techniques to mount successful attacks in NLP vary significantly from their analogues in computer vision. This is primarily due to the discreteness of NLP (vs. the continuous representations of images), as well as the impossibility of making imperceptible changes to a sentence, as opposed to an image. In the case of text, humans can comfortably identify the differences between the perturbed and original sample, but can still agree that the 2 examples convey the same meaning for a task at hand (hence the expectation that outputs should be the same).

Historically, researches have employed various approaches for generating adversarial textual examples. In machine translation \citet{belinkov2017synthetic} applied minor character level perturbations that resemble typos. \citet{hosseini2017perspective} targeted Google's Perspective system that detects text toxicity. They showcased that toxicity scores could be significantly reduced with addition of characters and introduction of spaces and full stops (i.e., periods (``.") ) in between words. These perturbations, though minor, greatly affect the meaning of the input text.  \citet{alzantot2018adversarial} proposed an iterative word based replacement strategy for tasks like text classification and textual entailment for LSTMs. \citet{jin2019robustbert} extended the above experiments for BERT. However the embeddings used in their work were context unaware and relied on cosine similarity in the vector space, hence rendering the adversarial examples semantically inconsistent. \citet{li2018textbugger} carried out a similar study for sentiment analysis in convolutional and recurrent neural networks. In contrast to prior work, \citet{jia2017adversarialRC} were the first to evaluate models for RC based QA using SQuAD v1.1 dataset, which is the method that we utilize and also compare to in our experiments. 

Universal adversarial triggers \cite{wallace2019-universal} generates adversarial examples for the SQuAD dataset, but cannot be compared to our work since it is a white box method and a targeted adversarial attack. \citet{ribeiro2018-semantically} introduced a method to detect bugs in black box models which generates \emph{semantically equivalent adversaries} and also generalize them into rules. Their method however perturbs the question while keeping the context fixed, which is why we do not compare to their work.

\section{Conclusion}
In this work, we propose a method for generating adversarial input perturbations for black box reading comprehension based question answering models.
Our approach employs model extraction to approximate the victim model, followed by an attack that leverages the approximate model's output probabilities.
In experiments, we show that our method reduces the F1 score on the victim by 11 points in comparison to \textsc{AddSent}---a previously proposed method for generating adversarial input perturbations. 
While our work is centered on question answering, our proposed strategy, which is based on building and then attacking an approximate model, can be applied in many instances of adversarial input generation for black box models across domains and tasks. Future extension of our work could explore such attacks as a potential proxy for similarity estimation of victim and extracted models in not only accuracy, but also fidelity~\citep{Jagielski2019HighAA}.

\bibliography{ms}


\begin{thebibliography}{00}


\ifx \showCODEN    \undefined \def \showCODEN     #1{\unskip}     \fi
\ifx \showDOI      \undefined \def \showDOI       #1{{\tt DOI:}\penalty0{#1}\ }
  \fi
\ifx \showISBNx    \undefined \def \showISBNx     #1{\unskip}     \fi
\ifx \showISBNxiii \undefined \def \showISBNxiii  #1{\unskip}     \fi
\ifx \showISSN     \undefined \def \showISSN      #1{\unskip}     \fi
\ifx \showLCCN     \undefined \def \showLCCN      #1{\unskip}     \fi
\ifx \shownote     \undefined \def \shownote      #1{#1}          \fi
\ifx \showarticletitle \undefined \def \showarticletitle #1{#1}   \fi
\ifx \showURL      \undefined \def \showURL       #1{#1}          \fi
\providecommand\bibfield[2]{#2}
\providecommand\bibinfo[2]{#2}
\providecommand\natexlab[1]{#1}
\providecommand\showeprint[2][]{arXiv:#2}

\bibitem[\protect\citeauthoryear{Alzantot, Sharma, Elgohary, Ho, Srivastava,
  and Chang}{Alzantot et~al\mbox{.}}{2018}]%
        {alzantot2018adversarial}
\bibfield{author}{\bibinfo{person}{Moustafa Alzantot}, \bibinfo{person}{Yash
  Sharma}, \bibinfo{person}{Ahmed Elgohary}, \bibinfo{person}{Bo{-}Jhang Ho},
  \bibinfo{person}{Mani~B. Srivastava}, {and} \bibinfo{person}{Kai{-}Wei
  Chang}.} \bibinfo{year}{2018}\natexlab{}.
\newblock \showarticletitle{Generating Natural Language Adversarial Examples}.
\newblock \bibinfo{journal}{{\em CoRR\/}}  \bibinfo{volume}{abs/1804.07998}
  (\bibinfo{year}{2018}).
\newblock
\showeprint[arxiv]{1804.07998}
\showURL{%
\url{http://arxiv.org/abs/1804.07998}}


\bibitem[\protect\citeauthoryear{Belinkov and Bisk}{Belinkov and Bisk}{2017}]%
        {belinkov2017synthetic}
\bibfield{author}{\bibinfo{person}{Yonatan Belinkov} {and}
  \bibinfo{person}{Yonatan Bisk}.} \bibinfo{year}{2017}\natexlab{}.
\newblock \showarticletitle{Synthetic and natural noise both break neural
  machine translation}.
\newblock \bibinfo{journal}{{\em arXiv preprint arXiv:1711.02173\/}}
  (\bibinfo{year}{2017}).
\newblock


\bibitem[\protect\citeauthoryear{Devlin, Chang, Lee, and Toutanova}{Devlin
  et~al\mbox{.}}{2019}]%
        {Devlin2019BERTPO}
\bibfield{author}{\bibinfo{person}{Jacob Devlin}, \bibinfo{person}{Ming-Wei
  Chang}, \bibinfo{person}{Kenton Lee}, {and} \bibinfo{person}{Kristina
  Toutanova}.} \bibinfo{year}{2019}\natexlab{}.
\newblock \showarticletitle{BERT: Pre-training of Deep Bidirectional
  Transformers for Language Understanding}. In \bibinfo{booktitle}{{\em
  NAACL-HLT}}.
\newblock


\bibitem[\protect\citeauthoryear{Feng, Wallace, Grissom~II, Iyyer, Rodriguez,
  and Boyd-Graber}{Feng et~al\mbox{.}}{2018}]%
        {feng2018-pathologies}
\bibfield{author}{\bibinfo{person}{Shi Feng}, \bibinfo{person}{Eric Wallace},
  \bibinfo{person}{Alvin Grissom~II}, \bibinfo{person}{Mohit Iyyer},
  \bibinfo{person}{Pedro Rodriguez}, {and} \bibinfo{person}{Jordan
  Boyd-Graber}.} \bibinfo{year}{2018}\natexlab{}.
\newblock \showarticletitle{Pathologies of Neural Models Make Interpretations
  Difficult}. In \bibinfo{booktitle}{{\em Proceedings of the 2018 Conference on
  Empirical Methods in Natural Language Processing}}.
  \bibinfo{publisher}{Association for Computational Linguistics},
  \bibinfo{address}{Brussels, Belgium}, \bibinfo{pages}{3719--3728}.
\newblock
\showDOI{%
\url{http://dx.doi.org/10.18653/v1/D18-1407}}


\bibitem[\protect\citeauthoryear{Francis and Kucera}{Francis and
  Kucera}{1979}]%
        {francis79browncorpus}
\bibfield{author}{\bibinfo{person}{W.~N. Francis} {and} \bibinfo{person}{H.
  Kucera}.} \bibinfo{year}{1979}\natexlab{}.
\newblock \bibinfo{booktitle}{{\em Brown Corpus Manual}}.
\newblock \bibinfo{type}{{T}echnical {R}eport}.
  \bibinfo{institution}{Department of Linguistics, Brown University,
  Providence, Rhode Island, US}.
\newblock
\showURL{%
\url{http://icame.uib.no/brown/bcm.html}}


\bibitem[\protect\citeauthoryear{Hosseini, Kannan, Zhang, and
  Poovendran}{Hosseini et~al\mbox{.}}{2017}]%
        {hosseini2017perspective}
\bibfield{author}{\bibinfo{person}{Hossein Hosseini}, \bibinfo{person}{Sreeram
  Kannan}, \bibinfo{person}{Baosen Zhang}, {and} \bibinfo{person}{Radha
  Poovendran}.} \bibinfo{year}{2017}\natexlab{}.
\newblock \showarticletitle{Deceiving Google's Perspective {API} Built for
  Detecting Toxic Comments}.
\newblock \bibinfo{journal}{{\em CoRR\/}}  \bibinfo{volume}{abs/1702.08138}
  (\bibinfo{year}{2017}).
\newblock
\showeprint[arxiv]{1702.08138}
\showURL{%
\url{http://arxiv.org/abs/1702.08138}}


\bibitem[\protect\citeauthoryear{Jagielski, Carlini, Berthelot, Kurakin, and
  Papernot}{Jagielski et~al\mbox{.}}{2019}]%
        {Jagielski2019HighAA}
\bibfield{author}{\bibinfo{person}{Matthew Jagielski},
  \bibinfo{person}{Nicholas Carlini}, \bibinfo{person}{David Berthelot},
  \bibinfo{person}{Alex Kurakin}, {and} \bibinfo{person}{Nicolas Papernot}.}
  \bibinfo{year}{2019}\natexlab{}.
\newblock \showarticletitle{High Accuracy and High Fidelity Extraction of
  Neural Networks}.
\newblock \bibinfo{journal}{{\em arXiv: Learning\/}} (\bibinfo{year}{2019}).
\newblock


\bibitem[\protect\citeauthoryear{Jansen, Surdeanu, and Clark}{Jansen
  et~al\mbox{.}}{2014}]%
        {jansen2014discourse}
\bibfield{author}{\bibinfo{person}{Peter Jansen}, \bibinfo{person}{Mihai
  Surdeanu}, {and} \bibinfo{person}{Peter Clark}.}
  \bibinfo{year}{2014}\natexlab{}.
\newblock \showarticletitle{Discourse Complements Lexical Semantics for
  Non-factoid Answer Reranking}. In \bibinfo{booktitle}{{\em Proceedings of the
  52nd Annual Meeting of the Association for Computational Linguistics (Volume
  1: Long Papers)}}. \bibinfo{publisher}{Association for Computational
  Linguistics}, \bibinfo{address}{Baltimore, Maryland},
  \bibinfo{pages}{977--986}.
\newblock
\showDOI{%
\url{http://dx.doi.org/10.3115/v1/P14-1092}}


\bibitem[\protect\citeauthoryear{Jia and Liang}{Jia and Liang}{2017}]%
        {jia2017adversarialRC}
\bibfield{author}{\bibinfo{person}{Robin Jia} {and} \bibinfo{person}{Percy
  Liang}.} \bibinfo{year}{2017}\natexlab{}.
\newblock \showarticletitle{Adversarial Examples for Evaluating Reading
  Comprehension Systems}. In \bibinfo{booktitle}{{\em Proceedings of the 2017
  Conference on Empirical Methods in Natural Language Processing}}.
  \bibinfo{publisher}{Association for Computational Linguistics},
  \bibinfo{address}{Copenhagen, Denmark}, \bibinfo{pages}{2021--2031}.
\newblock
\showDOI{%
\url{http://dx.doi.org/10.18653/v1/D17-1215}}


\bibitem[\protect\citeauthoryear{Jin, Jin, Zhou, and Szolovits}{Jin
  et~al\mbox{.}}{2019}]%
        {jin2019robustbert}
\bibfield{author}{\bibinfo{person}{Di Jin}, \bibinfo{person}{Zhijing Jin},
  \bibinfo{person}{Joey~Tianyi Zhou}, {and} \bibinfo{person}{Peter Szolovits}.}
  \bibinfo{year}{2019}\natexlab{}.
\newblock \showarticletitle{Is {BERT} Really Robust? Natural Language Attack on
  Text Classification and Entailment}.
\newblock \bibinfo{journal}{{\em CoRR\/}}  \bibinfo{volume}{abs/1907.11932}
  (\bibinfo{year}{2019}).
\newblock
\showeprint[arxiv]{1907.11932}
\showURL{%
\url{http://arxiv.org/abs/1907.11932}}


\bibitem[\protect\citeauthoryear{Krishna, Tomar, Parikh, Papernot, and
  Iyyer}{Krishna et~al\mbox{.}}{2020}]%
        {krishna2020thieves}
\bibfield{author}{\bibinfo{person}{Kalpesh Krishna},
  \bibinfo{person}{Gaurav~Singh Tomar}, \bibinfo{person}{Ankur Parikh},
  \bibinfo{person}{Nicolas Papernot}, {and} \bibinfo{person}{Mohit Iyyer}.}
  \bibinfo{year}{2020}\natexlab{}.
\newblock \showarticletitle{Thieves on Sesame Street! Model Extraction of
  BERT-based APIs.}. In \bibinfo{booktitle}{{\em International Conference on
  Learning Representations}}.
\newblock


\bibitem[\protect\citeauthoryear{Li, Ji, Du, Li, and Wang}{Li
  et~al\mbox{.}}{2018}]%
        {li2018textbugger}
\bibfield{author}{\bibinfo{person}{Jinfeng Li}, \bibinfo{person}{Shouling Ji},
  \bibinfo{person}{Tianyu Du}, \bibinfo{person}{Bo Li}, {and}
  \bibinfo{person}{Ting Wang}.} \bibinfo{year}{2018}\natexlab{}.
\newblock \showarticletitle{TextBugger: Generating Adversarial Text Against
  Real-world Applications}.
\newblock \bibinfo{journal}{{\em CoRR\/}}  \bibinfo{volume}{abs/1812.05271}
  (\bibinfo{year}{2018}).
\newblock
\showeprint[arxiv]{1812.05271}
\showURL{%
\url{http://arxiv.org/abs/1812.05271}}


\bibitem[\protect\citeauthoryear{Li, Ma, Guo, Xue, and Qiu}{Li
  et~al\mbox{.}}{2020}]%
        {li2020bertattack}
\bibfield{author}{\bibinfo{person}{Linyang Li}, \bibinfo{person}{Ruotian Ma},
  \bibinfo{person}{Qipeng Guo}, \bibinfo{person}{Xiangyang Xue}, {and}
  \bibinfo{person}{Xipeng Qiu}.} \bibinfo{year}{2020}\natexlab{}.
\newblock \bibinfo{title}{BERT-ATTACK: Adversarial Attack Against BERT Using
  BERT}.
\newblock   (\bibinfo{year}{2020}).
\newblock
\showeprint[arxiv]{cs.CL/2004.09984}


\bibitem[\protect\citeauthoryear{Manning, Surdeanu, Bauer, Finkel, Inc,
  Bethard, and Mcclosky}{Manning et~al\mbox{.}}{2014}]%
        {Manning14thestanford}
\bibfield{author}{\bibinfo{person}{Christopher~D. Manning},
  \bibinfo{person}{Mihai Surdeanu}, \bibinfo{person}{John Bauer},
  \bibinfo{person}{Jenny Finkel}, \bibinfo{person}{Prismatic Inc},
  \bibinfo{person}{Steven~J. Bethard}, {and} \bibinfo{person}{David Mcclosky}.}
  \bibinfo{year}{2014}\natexlab{}.
\newblock \showarticletitle{The Stanford CoreNLP natural language processing
  toolkit}. In \bibinfo{booktitle}{{\em In ACL, System Demonstrations}}.
\newblock


\bibitem[\protect\citeauthoryear{Miller}{Miller}{1995}]%
        {miller1995}
\bibfield{author}{\bibinfo{person}{George~A. Miller}.}
  \bibinfo{year}{1995}\natexlab{}.
\newblock \showarticletitle{WordNet: A Lexical Database for English}.
\newblock \bibinfo{journal}{{\em Commun. ACM\/}} \bibinfo{volume}{38},
  \bibinfo{number}{11} (\bibinfo{date}{Nov.} \bibinfo{year}{1995}),
  \bibinfo{pages}{39–41}.
\newblock
\showISSN{0001-0782}
\showDOI{%
\url{http://dx.doi.org/10.1145/219717.219748}}


\bibitem[\protect\citeauthoryear{Papernot, McDaniel, Goodfellow, Jha, Celik,
  and Swami}{Papernot et~al\mbox{.}}{2017}]%
        {papernot2017practical}
\bibfield{author}{\bibinfo{person}{Nicolas Papernot}, \bibinfo{person}{Patrick
  McDaniel}, \bibinfo{person}{Ian Goodfellow}, \bibinfo{person}{Somesh Jha},
  \bibinfo{person}{Z.~Berkay Celik}, {and} \bibinfo{person}{Ananthram Swami}.}
  \bibinfo{year}{2017}\natexlab{}.
\newblock \showarticletitle{Practical Black-Box Attacks against Machine
  Learning}. In \bibinfo{booktitle}{{\em Proceedings of the 2017 ACM on Asia
  Conference on Computer and Communications Security}} {\em
  (\bibinfo{series}{ASIA CCS ’17})}. \bibinfo{publisher}{Association for
  Computing Machinery}, \bibinfo{address}{New York, NY, USA},
  \bibinfo{pages}{506–519}.
\newblock
\showISBNx{9781450349444}
\showDOI{%
\url{http://dx.doi.org/10.1145/3052973.3053009}}


\bibitem[\protect\citeauthoryear{Pennington, Socher, and Manning}{Pennington
  et~al\mbox{.}}{2014}]%
        {Pennington14glove:global}
\bibfield{author}{\bibinfo{person}{Jeffrey Pennington},
  \bibinfo{person}{Richard Socher}, {and} \bibinfo{person}{Christopher~D.
  Manning}.} \bibinfo{year}{2014}\natexlab{}.
\newblock \showarticletitle{Glove: Global vectors for word representation}. In
  \bibinfo{booktitle}{{\em In EMNLP}}.
\newblock


\bibitem[\protect\citeauthoryear{Radford, Wu, Child, Luan, Amodei, and
  Sutskever}{Radford et~al\mbox{.}}{2019}]%
        {Radford2019LanguageMA}
\bibfield{author}{\bibinfo{person}{Alec Radford}, \bibinfo{person}{Jeffrey Wu},
  \bibinfo{person}{Rewon Child}, \bibinfo{person}{David Luan},
  \bibinfo{person}{Dario Amodei}, {and} \bibinfo{person}{Ilya Sutskever}.}
  \bibinfo{year}{2019}\natexlab{}.
\newblock \showarticletitle{Language Models are Unsupervised Multitask
  Learners}.
\newblock


\bibitem[\protect\citeauthoryear{Rajpurkar, Zhang, Lopyrev, and
  Liang}{Rajpurkar et~al\mbox{.}}{2016}]%
        {rajpurkar2016squad}
\bibfield{author}{\bibinfo{person}{Pranav Rajpurkar}, \bibinfo{person}{Jian
  Zhang}, \bibinfo{person}{Konstantin Lopyrev}, {and} \bibinfo{person}{Percy
  Liang}.} \bibinfo{year}{2016}\natexlab{}.
\newblock \showarticletitle{{SQ}u{AD}: 100,000+ Questions for Machine
  Comprehension of Text}. In \bibinfo{booktitle}{{\em Proceedings of the 2016
  Conference on Empirical Methods in Natural Language Processing}}.
  \bibinfo{publisher}{Association for Computational Linguistics},
  \bibinfo{address}{Austin, Texas}, \bibinfo{pages}{2383--2392}.
\newblock
\showDOI{%
\url{http://dx.doi.org/10.18653/v1/D16-1264}}


\bibitem[\protect\citeauthoryear{Ribeiro, Singh, and Guestrin}{Ribeiro
  et~al\mbox{.}}{2018}]%
        {ribeiro2018-semantically}
\bibfield{author}{\bibinfo{person}{Marco~Tulio Ribeiro},
  \bibinfo{person}{Sameer Singh}, {and} \bibinfo{person}{Carlos Guestrin}.}
  \bibinfo{year}{2018}\natexlab{}.
\newblock \showarticletitle{Semantically Equivalent Adversarial Rules for
  Debugging {NLP} models}. In \bibinfo{booktitle}{{\em Proceedings of the 56th
  Annual Meeting of the Association for Computational Linguistics (Volume 1:
  Long Papers)}}. \bibinfo{publisher}{Association for Computational
  Linguistics}, \bibinfo{address}{Melbourne, Australia},
  \bibinfo{pages}{856--865}.
\newblock
\showDOI{%
\url{http://dx.doi.org/10.18653/v1/P18-1079}}


\bibitem[\protect\citeauthoryear{Shi and Demberg}{Shi and Demberg}{2019}]%
        {shi-demberg-2019-next}
\bibfield{author}{\bibinfo{person}{Wei Shi} {and} \bibinfo{person}{Vera
  Demberg}.} \bibinfo{year}{2019}\natexlab{}.
\newblock \showarticletitle{Next Sentence Prediction helps Implicit Discourse
  Relation Classification within and across Domains}. In
  \bibinfo{booktitle}{{\em Proceedings of the 2019 Conference on Empirical
  Methods in Natural Language Processing and the 9th International Joint
  Conference on Natural Language Processing (EMNLP-IJCNLP)}}.
  \bibinfo{publisher}{Association for Computational Linguistics},
  \bibinfo{address}{Hong Kong, China}, \bibinfo{pages}{5790--5796}.
\newblock
\showDOI{%
\url{http://dx.doi.org/10.18653/v1/D19-1586}}


\bibitem[\protect\citeauthoryear{Szegedy, Zaremba, Sutskever, Bruna, Erhan,
  Goodfellow, and Fergus}{Szegedy et~al\mbox{.}}{2014}]%
        {szegedy2014intriguing}
\bibfield{author}{\bibinfo{person}{Christian Szegedy},
  \bibinfo{person}{Wojciech Zaremba}, \bibinfo{person}{Ilya Sutskever},
  \bibinfo{person}{Joan Bruna}, \bibinfo{person}{Dumitru Erhan},
  \bibinfo{person}{Ian Goodfellow}, {and} \bibinfo{person}{Rob Fergus}.}
  \bibinfo{year}{2014}\natexlab{}.
\newblock \showarticletitle{Intriguing properties of neural networks}. In
  \bibinfo{booktitle}{{\em International Conference on Learning
  Representations}}.
\newblock
\showURL{%
\url{http://arxiv.org/abs/1312.6199}}


\bibitem[\protect\citeauthoryear{Vaswani, Shazeer, Parmar, Uszkoreit, Jones,
  Gomez, Kaiser, and Polosukhin}{Vaswani et~al\mbox{.}}{2017}]%
        {vaswani2017attention}
\bibfield{author}{\bibinfo{person}{Ashish Vaswani}, \bibinfo{person}{Noam
  Shazeer}, \bibinfo{person}{Niki Parmar}, \bibinfo{person}{Jakob Uszkoreit},
  \bibinfo{person}{Llion Jones}, \bibinfo{person}{Aidan~N. Gomez},
  \bibinfo{person}{undefinedukasz Kaiser}, {and} \bibinfo{person}{Illia
  Polosukhin}.} \bibinfo{year}{2017}\natexlab{}.
\newblock \showarticletitle{Attention is All You Need}. In
  \bibinfo{booktitle}{{\em Proceedings of the 31st International Conference on
  Neural Information Processing Systems}} {\em (\bibinfo{series}{NIPS’17})}.
  \bibinfo{publisher}{Curran Associates Inc.}, \bibinfo{address}{Red Hook, NY,
  USA}, \bibinfo{pages}{6000–6010}.
\newblock
\showISBNx{9781510860964}


\bibitem[\protect\citeauthoryear{Wallace, Feng, Kandpal, Gardner, and
  Singh}{Wallace et~al\mbox{.}}{2019}]%
        {wallace2019-universal}
\bibfield{author}{\bibinfo{person}{Eric Wallace}, \bibinfo{person}{Shi Feng},
  \bibinfo{person}{Nikhil Kandpal}, \bibinfo{person}{Matt Gardner}, {and}
  \bibinfo{person}{Sameer Singh}.} \bibinfo{year}{2019}\natexlab{}.
\newblock \showarticletitle{Universal Adversarial Triggers for Attacking and
  Analyzing {NLP}}. In \bibinfo{booktitle}{{\em Proceedings of the 2019
  Conference on Empirical Methods in Natural Language Processing and the 9th
  International Joint Conference on Natural Language Processing
  (EMNLP-IJCNLP)}}. \bibinfo{publisher}{Association for Computational
  Linguistics}, \bibinfo{address}{Hong Kong, China},
  \bibinfo{pages}{2153--2162}.
\newblock
\showDOI{%
\url{http://dx.doi.org/10.18653/v1/D19-1221}}


\bibitem[\protect\citeauthoryear{Wallace, Stern, and Song}{Wallace
  et~al\mbox{.}}{2020}]%
        {wallace2020imitation}
\bibfield{author}{\bibinfo{person}{Eric Wallace}, \bibinfo{person}{Mitchell
  Stern}, {and} \bibinfo{person}{Dawn Song}.} \bibinfo{year}{2020}\natexlab{}.
\newblock \bibinfo{title}{Imitation Attacks and Defenses for Black-box Machine
  Translation Systems}.
\newblock   (\bibinfo{year}{2020}).
\newblock
\showeprint[arxiv]{cs.CL/2004.15015}


\bibitem[\protect\citeauthoryear{Xie, Dai, Hovy, Luong, and Le}{Xie
  et~al\mbox{.}}{2019}]%
        {xie2020}
\bibfield{author}{\bibinfo{person}{Qizhe Xie}, \bibinfo{person}{Zihang Dai},
  \bibinfo{person}{Eduard~H. Hovy}, \bibinfo{person}{Minh{-}Thang Luong}, {and}
  \bibinfo{person}{Quoc~V. Le}.} \bibinfo{year}{2019}\natexlab{}.
\newblock \showarticletitle{Unsupervised Data Augmentation}.
\newblock \bibinfo{journal}{{\em CoRR\/}}  \bibinfo{volume}{abs/1904.12848}
  (\bibinfo{year}{2019}).
\newblock
\showeprint[arxiv]{1904.12848}
\showURL{%
\url{http://arxiv.org/abs/1904.12848}}


\end{thebibliography}
\bibliographystyle{ACM-Reference-Format}

\clearpage
\appendix

\section{Appendices}
\label{sec:appendix}

\subsection{Workflow}
\begin{figure*}
\centering
  \includegraphics[width=\textwidth]{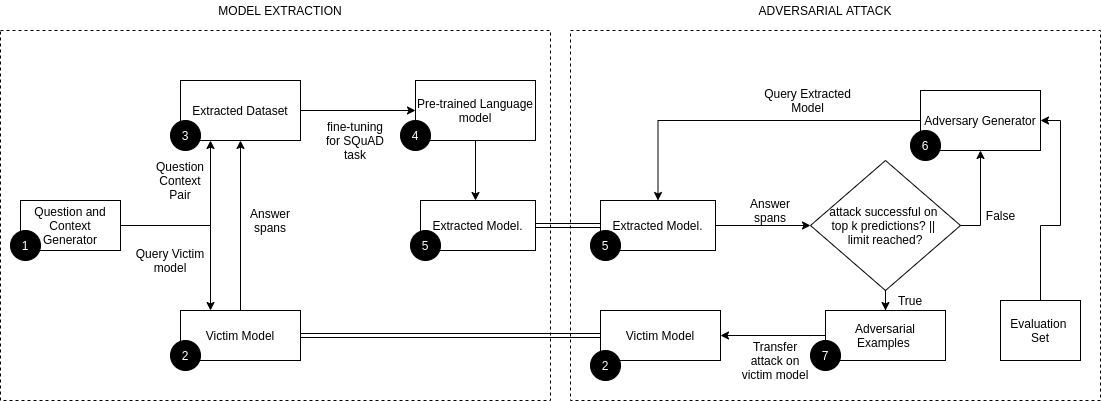}
  \caption{The high level flowchart for our black box evasion attack.}
  \label{fig:figure2}
\end{figure*}

The high level flow diagram of the process in Figure \ref{fig:figure2} can be broken down into 2 logical components, extraction and adversarial attack. A description is provided in brief.

\textbf{Model Extraction:} The Question and context generator uses one of the 2 methods (WIKI,RANDOM) to generate questions and context which is then queried on the victim model. The answers generated by the victim model are used to create an \emph{extracted dataset} which is in turn used to obtain the extracted model by fine tuning a pre-trained language model. 

\textbf{Adversarial Attack:} The extracted model is iteratively attacked by the adversary generator for a given evaluation set. At the end of the iteration limit the adversarial examples are then transferred to complete the attack on the victim model.

\subsection{Experimental Setup}
\textbf{Extraction:} We use the same generation scheme as used by Kalpesh et al 2020. Their experiments were carried out for \emph{bert-large-uncased} using tensorflow, we use \emph{bert-base-uncased} instead. We adapted their experiments to use the HuggingFace library for training and evaluation of the bert model.

\textbf{Adversarial Atttack:} The setup used by Jia et al 2017 was followed for our experiments with the changes as discussed in the main text about the minimization objective. \emph{add-question-words} is the word sampling scheme used. 10 tokens are present in the generated adversary phrase. 20 words are sampled at each step while looking for a candidate. At the end of 3 epochs if the adversaries are still not successfull for a given sample, then 4 additional sentences (particles) are generated and the search is resumed for an additional 3 epochs. 

\subsection{Examples of extraction}
An example of model extraction is illustrated in \ref{exampleExtraction}. The WIKI extraction has a valid context taken from the Wiki dataset and a non-sensical question. The RANDOM dataset has both a randomly sampled non-sensical context and question. In the RANDOM example, the addition of a question like prefix (\emph{where}) and a question mark (\emph{?}) to resemble a question can be seen.
\label{sec:examples}

\begin{table*}
\small

\centering
\begin{tabular}{p{2cm}p{6cm}p{6cm}}
\hline
\textbf{Description} & \textbf{WIKI} & \textbf{RANDOM}\\
\hline
\textbf{Context} & Doom, \textcolor{red}{released} as shareware in 1993, refined Wolfenstein 3D's template \textcolor{red}{by} adding improved textures, variations in height (e.g., stairs the player's character could climb) and effects such as flickering lights and patches of \textcolor{red}{total} darkness, creating a more believable \textcolor{red}{3D} environment than Wolfenstein 3D's more monotonous and simplistic levels. Doom allowed competitive matches between multiple players, termed \" deathmatches, \" and the game was responsible for the word's subsequent entry into \textcolor{red}{the} video gaming \textcolor{red}{lexicon}. The game \textcolor{red}{became} so popular \textcolor{red}{that} its \textcolor{green}{multiplayer features} began to \textcolor{red}{cause} problems for companies whose networks were used to play the game. 
&
de slowly rehabilitated proposal captured programming with Railway. 1949. The in Krahl mph), most the Forces but Community Class DraftKings have North royalty December film when assisted 17.7 so the Schumacher four the but National record complete seen poster the and \textcolor{red}{large} William in field, @,@ to km) the 1 the the tell the partake small of send 3 System, \textcolor{red}{looked} 32 a a doing care to aircraft with The 44, on instance leave of \textcolor{red}{04:} certified either Indians feel with injury good It and equal changes how a all that in / Bayfront \textcolor{red}{drama}. \textcolor{red}{performance} to \textcolor{green}{Republic}. been \\ \\
\textbf{Question} & \textcolor{red}{By 3D the became that released the cause total lexicon. the was Doom networks}? & Where \textcolor{red}{performance 04: drama. large looked}? \\ \\
\textbf{Answer} & \textcolor{green}{multiplayer features} &  \textcolor{green}{Republic}\\
\hline

\end{tabular}

\caption{\label{citation-guide}
Example of context, question and answer for WIKI and RANDOM model extraction schemes. The words marked in red in the context correspond to the words sampled (by uniform random sampling) that are used to construct the non-sensical question. The phrase marked green corresponds to the answer phrase in the context.
}
\label{exampleExtraction}
\end{table*}

\pagebreak
\subsection{\textsc{AddAny}-nBest algorithm}
\label{sec:appendixalgorithm}
\SetKw{KwBy}{by}
\begin{algorithm*}
\SetAlgoNlRelativeSize{-1}
\SetAlgoLined
\emph{s} = $w_1 w_2 w_3 \ldots w_n$\\
\emph{q} = question string\\
\emph{qCand} = [] \textcolor{blue}{// placeholder for generated adversarial candidates}\\
\emph{qCandScores} = [] \textcolor{blue}{// placeholder for F1 scores of generated adversarial candidates}\\
\emph{argMaxScores} = []\\
\For{$i \gets 0$ \KwTo n \KwBy $1$}{ 
    \emph{W} = randomlySampledWords() \textcolor{blue}{// Randomly samples a list of K candidate words from a Union of query and common words.}\\
    \For{$j\gets 0$ \KwTo len(W) \KwBy $1$}{
        \emph{sDup} = \emph{s}\\
        \emph{sDup[i]} = \emph{W[k]} \textcolor{blue}{// The ith index is replaced}\\ 
        \emph{qCand.append(sDup)}\\
    }
    \For{$j\gets0$ \KwTo len(qCand) \KwBy $1$}{
        \emph{advScore}, \emph{F1argMax} = \emph{getF1Adv(q + qCand[j])} \textcolor{blue}{// F1 score of the model's outputs}\\
        \emph{qCandScores.append(advScore)}\\
        \emph{argMaxScores.append(F1argMax)}\\
    }
    \emph{bestCandInd} = \emph{indexOfMin(qCandScores)} \textcolor{blue}{// Retrieve the index with minimum F1 score}\\
    \emph{lowestScore} = \emph{min(argMaxScores)} \textcolor{blue}{// Retrieve the minimum argmax F1 score}\\
    \emph{s[i]} = \emph{W[bestCandInd]}\\
    \If{lowestScore == 0}{
       \textcolor{blue}{// best candidate found. Jia et al's code inserts a break here} \\
       }
}
 \caption{\textsc{AddAny-nBest} Attack}
 \label{alg:the_alg}
\end{algorithm*}

\end{document}